\documentclass[conference]{IEEEtran}
\usepackage{cite}
\usepackage{amsmath,amssymb,amsfonts,mathtools}
\usepackage{algorithmic}
\usepackage{graphicx}
\usepackage{textcomp}
\usepackage{xcolor}
\usepackage{subfigure}
\def\BibTeX{{\rm B\kern-.05em{\sc i\kern-.025em b}\kern-.08em
    T\kern-.1667em\lower.7ex\hbox{E}\kern-.125emX}}
\begin{document}

\title{Neural Networks Versus Conventional Filters for Inertial-Sensor-based Attitude Estimation}

\author{\IEEEauthorblockN{Daniel Weber}
\IEEEauthorblockA{\textit{Measurement and Diagnostic Technology} \\
\textit{Technische Universität Berlin}\\
Berlin, Germany \\
d.weber.1@tu-berlin.de}
\and
\IEEEauthorblockN{Clemens Gühmann}
\IEEEauthorblockA{\textit{Measurement and Diagnostic Technology} \\
\textit{Technische Universität Berlin}\\
Berlin, Germany \\
clemens.guehmann@tu-berlin.de}
\and
\IEEEauthorblockN{Thomas Seel}
\IEEEauthorblockA{\textit{Control Systems Group} \\
\textit{Technische Universität Berlin}\\
Berlin, Germany \\
thomas.seel@tu-berlin.de}
}

\maketitle
\begin{abstract}

Inertial measurement units are commonly used to estimate the attitude of moving objects. 
Numerous nonlinear filter approaches have been proposed for solving the inherent sensor fusion problem. However, when a large range of different dynamic and static rotational and translational motions is considered, the attainable accuracy is limited by the need for situation-dependent adjustment of accelerometer and gyroscope fusion weights. 
We investigate to what extent these limitations can be overcome by means of artificial neural networks and how much domain-specific optimization of the neural network model is required to outperform the conventional filter solution. A diverse set of motion recordings with a marker-based optical ground truth is used for performance evaluation and comparison.
The proposed neural networks are found to outperform the conventional filter across all motions only if domain-specific optimizations are introduced. We conclude that they are a promising tool for inertial-sensor-based real-time attitude estimation, but both expert knowledge and rich datasets are required to achieve top performance.

\end{abstract}


\section{Introduction}
Inertial sensors have been used for several decades in aerospace system for attitude control and navigation. Drastic advances in microelectromechanic systems (MEMS) have led to the development of miniaturized strapdown inertial measurement units (IMUs), which entered a multitude of new application domains from autonomous drones to ambulatory human motion tracking. 

\begin{figure}
    \centering
    \includegraphics[width=0.45\textwidth]{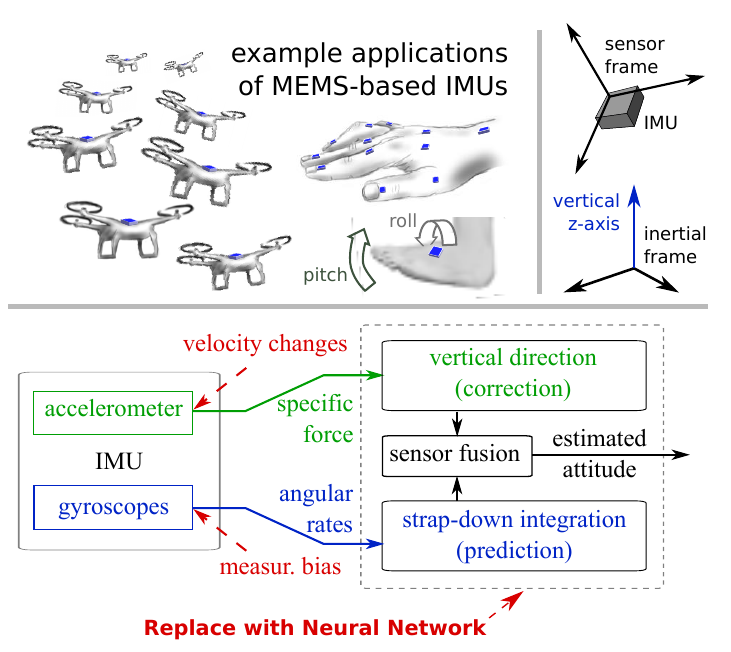}
    \caption{Sensor fusion workflow in attitude estimation. In many applications, IMUs are used to estimate orientations for the horizontal plane in real-time. Commonly used sensor fusion filters might be outperformed by sufficiently advanced and well-trained neural networks. Graphic based on \cite{beuchert_overcoming_2020}.}
    \label{fig:attitude_workflow}
\end{figure}

In strapdown IMUs, the angular rate and acceleration -- and often also the magnetic field vector -- are measured in a sensor-intrinsic three-dimensional coordinate system, which moves along with the sensor. Estimating the orientation, velocity or position of the sensor with respect to some inertial frame requires strapdown integration of the angular rates and sensor fusion of the aforementioned raw measurement signals (cf. Figure~\ref{fig:attitude_workflow}).

To estimate the orientation of an IMU from its raw measurement signals in real-time is a fundamental standard problem of inertial sensor fusion. A large variety of filter algorithms have been proposed previously, some of which are implemented in motion processing units of modern miniature IMUs. It is well known that the attitude of the sensor can be determined by 6D sensor fusion, i.e. fusing 3D gyroscope and 3D accelerometer readings, while estimating the full orientation (attitude and heading) requires 9D sensor fusion, i.e. using 3D magnetometer readings in addition to the 6D signals.

Existing solutions to inertial attitude estimation are typically model-based and heuristically parameterized. They use mathematical models of measurement errors and three-dimensional rotations and transformations of the gravitational acceleration. They require a reasonable choice of covariance matrices, fusion weights or parameters that define how weights are adjusted. While considerably high accuracies have been achieved with such solution approaches in many application domains, it is also well-known that different parameterizations perform differently well for different types of motions and disturbances. In fact, to the best of our knowledge, there is to date no filter algorithm that yields consistently small errors across all types of motion that a MEMS-based IMU might perform.

Abundant research has demonstrated the capabilities of artificial neural networks in providing data-based solutions to problems that have conventionally been addressed by model-based approaches. If sufficiently large amounts of data and computation capability are available, generally usable solutions may be found for most problems. While ample work has shown that a number of problems can \emph{also} be solved using neural networks, the practically more relevant question whether neural networks can outperform conventional solutions often remains unanswered.

In the present work, we investigate whether a neural network can solve the real-time attitude estimation task with similar or even better performance than a state-of-the-art inertial orientation estimation filter. Moreover, we analyze at which cost this can be achieved, in terms of required number of datasets, required complexity and application-specific structure of the neural network.

\section{Related Work}

We first briefly review the state-of-the-art in real-time attitude estimation from inertial sensor signals and then describe previous work on the use of artificial neural networks for inertial motion analysis.

\subsection{Inertial Attitude Estimation}

As mentioned above, the attitude of an IMU can be determined by sensor fusion of the accelerometer and gyroscope readings. Accelerometers yield accurate attitude information in static conditions, i.e. when the sensor moves with constant velocity. Under dynamic conditions, however, their readings are only useful under certain assumption, for example that the average change of velocity is zero on sufficiently large time scales. Gyroscopes yield highly accurate information on the change of attitude. However, pure strapdown integration of the angular rates is prone to drift resulting from measurement bias, noise, clipping and undersampling. Accurate attitude estimation under non-static conditions requires sensor fusion of both 3D signals.

A number of different solutions have been proposed for this task. Categorizations and comparisons of different algorithms can be found, for example, in \cite{caruso_accuracy_2019,ricci_orientation_2016}. Most filters use either an extended Kalman filter scheme or a complementary filter scheme, and unit quaternions are a common choice for mathematical representation of the three-dimensional orientation. The balance between gyroscope-based strapdown integration and accelerometer-based drift correction is typically adjusted to the specific application by manual tuning of covariance matrices or other fusion weights. Methods have been proposed that analyze the accelerometer norm to distinguish static and dynamic motion phases and adjust the fusion weights in real time.

A rather recently developed quaternion-based orientation estimation filter is described in \cite{seel_eliminating_2017}. It uses geodetic accelerometer-based correction steps and optional magnetometer-based correction steps for heading estimation. The correction steps are parametrized by intuitively interpretable time constants, which are adjusted automatically if the accelerometer norm is far from the static value or has been close to that value for several consecutive time steps. The performance of this filter and five other state-of-the-art filters has recently been evaluated across a wide range of motions. For all filters, errors between two and five degrees were found for different speeds of motion \cite{caruso_comparison_2020}. To the best of our knowledge, a significantly more accurate solution for attitude estimation in MEMS-based IMUs does not exist.

\subsection{Neural Networks for Attitude Estimation}
In inertial motion tracking, neural networks have mostly been applied to augment existing conventional filter solutions. In \cite{brossard_rins-w_2020} a  Recurrent Neural Network (RNN) is used for movement detection to decide which Kalman filter should be applied to the current system state. In \cite{chiang_artificial_2009} a feed forward neural network is used as for smoothing the output of a Kalman filter, while an RNN is used for data pre-processing of Kalman filter inputs in \cite{rambach_learning_2016}. A similar approach is used in \cite{brossard_denoising_2020}, where a convolutional neural network is used for error correction of the gyroscope signal as part of a strapdown integration. 

In \cite{esfahani_aboldeepio_2019} and \cite{esfahani_orinet_2020} RNNs are used as blackboxes for the orientation integration over time. While the former uses a combination of gyroscope and visual data, the latter only relies on the gyroscope achieving similar results. In a few more recent works, neural networks have been applied directly as blackboxes for angle estimation problems. In \cite{zimmermann_imu--segment_2018} an RNN is used for human limb assignment and orientation estimation of IMUs that are attached to human limbs. It achieved a high accuracy at the assignment problem but was only partially successful at the orientation estimation problem. In \cite{chen_deep_2019} a bidirectional RNN is used for velocity and heading estimation on a two-dimensional plane in polar coordinates.

To conclude, an end-to-end neural network model for IMU-based attitude estimation has not been developed yet. All of the presented neural networks are either an addition to classical filters for attitude estimation or they address different problems.

\section{Problem Statement}

Consider an inertial sensor with an intrinsic right-handed coordinate system $\mathcal{S}$. Neglect the rotation of the Earth and define an inertial frame of reference $\mathcal{E}$ with vertical z-axis. The orientation of the sensor with respect to the reference frame is then described by the rotation between both coordinate systems, which can be expressed as a unit quaternion, a rotation matrix, a set of three Euler angles or a single angle and a corresponding rotation axis. Both frames are said to have the same attitude if the axis of that rotation is vertical. 

If the true orientation of the sensor is given by the unit quaternion 
$\mathbf{q}$ and an attitude estimation algorithm yields an estimate
$\mathbf{\hat{q}}$, then $\mathbf{q}_\text{err}=\mathbf{q}\otimes\mathbf{\hat{q}}^{-1}$ is the estimation error quaternion expressed in reference frame axes. The attitude estimation is said to be perfect if the estimated orientation is correct up to a rotation around the vertical axis. This is the case if the rotation axis of $\mathbf{q}_\text{err}$ is vertical. If that axis is not vertical, then $\mathbf{q}_\text{err}$ can be decomposed into a rotation $\mathbf{q}_\text{head\,err}$ around the vertical axis and a rotation $\mathbf{q}_\text{att\,err}$ around a horizontal axis. For any given $\mathbf{q}_\text{err}$ with real part $w_\text{err}$ and third imaginary part $z_\text{err}$, the smallest possible rotation angle of $\mathbf{q}_\text{att\,err}$ is $2\arccos\sqrt{w_\text{err}^2+z_\text{err}^2}$. This corresponds to the smallest rotation by which one would need to correct the estimate $\mathbf{\hat{q}}$ to make its attitude error zero in the aforementioned sense.

These definitions allow us to formulate the following attitude estimation problem: Given a sampled sequence of three-dimensional accelerometer and gyroscope readings of a MEMS-based IMU moving freely in three-dimensional space, estimate the attitude of that IMU with respect to the reference frame at each sampling instant only based on current and previous samples. Denote the sensor readings by $\mathbf{a}(t)$ and $\mathbf{g}(t)$, respectively, with $t=T_s,2T_s,3T_s,...,NT_s$ being the discrete time and $N$ the number of samples. The desired algorithm should then yield a sampled sequence of estimates $\mathbf{\hat{q}}(t)$ with a possibly small cumulative attitude estimation error $e_{\alpha,\text{RMS}}$ defined by
\begin{gather}
  \left[w_\text{err}(t),x_\text{err}(t),y_\text{err}(t),z_\text{err}(t)\right]^\intercal = \mathbf{q}(t)\otimes\left(\mathbf{\hat{q}}(t)\right)^{-1},\\
  e_\alpha(t) =  2\arccos\left(\sqrt{w_\text{err}(t)^2+z_\text{err}(t)^2}\right),\\
  e_{\alpha,\text{RMS}} = \sqrt{\frac{1}{N}\sum_{t=T_s}^{NT_s}\left.e_\alpha(t)\right.^2},
\end{gather}
where $\mathbf{q}(t)$ is the true orientation of the sensor at time $t$. In the following sections, we aim to develop an artificial neural network that solves the given problem and compare it to established attitude estimation filters.

\section{Neural Network Model}
In this work, a neural network model with state-of-the-art best practices for time series will be implemented. Building upon that, further optimizations are introduced that utilize domain-specific knowledge.

\subsection{Neural Network Structure with general best practices}
The performance of a neural network model depends on the model architecture and the training process. First, we identify potential model architectures for attitude estimation. After that, we develop an optimized training process for these architectures.
\begin{figure}
    \centering
    \includegraphics[width=0.45\textwidth]{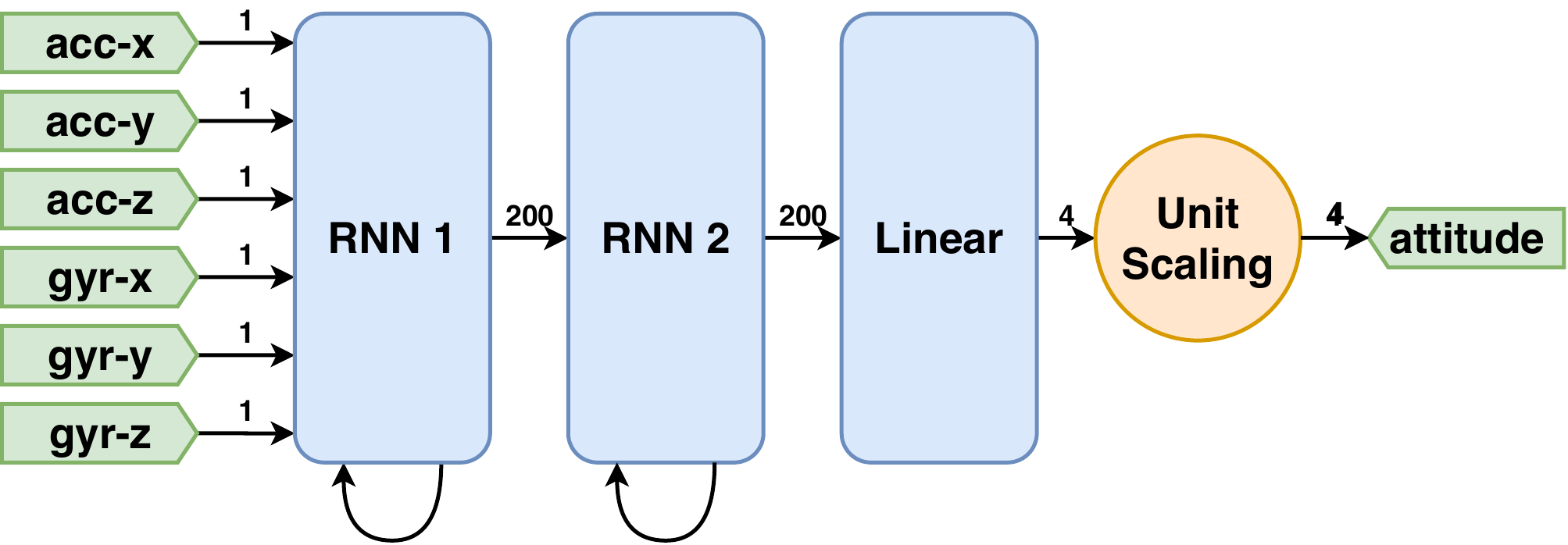}
    \caption{RNN model for attitude estimation. The measured signals are processed by 2 RNN layers with 200 neurons followed by a linear layer with 4 neurons, which, scaled to unit length, represent the attitude quaternion.}
    \label{fig:model_rnn}
\end{figure}

The model architecture consists of multiple layers that may be connected in multiple ways leading to different characteristics. First, a method for modelling the dynamic system states has to be chosen.  A common practice is to connect the model output to the model input creating an autoregressive model that stores the system state information in the single autoregressive connection. For longer sequences, the autoregressive model's inherent sequential nature prevents parallelization and therefore an efficient use of hardware acceleration, which slows down the training. Using neural network layers that are able to model system states avoids the need of autoregression for dynamic systems. The most commonly used ones are Recurrent Neural Networks (RNNs) and Temporal Convolutional Networks (TCNs).

RNNs have recurrent connections between samples in their hidden activations for modelling the state of a dynamic system. There are different variants of RNNs with Long Short-Term Memories (LSTMs) being the most prevalent \cite{gonzalez_non-linear_2018}. LSTMs add three gates to the classical RNN, which regulate the information flow of its hidden activations. This stabilizes the stored state, enabling the application to systems with long-term dependencies, like integrating movements over a long amount of time. Because LSTMs are prone to overfitting, several regularization methods for sequential neural networks have been developed \cite{merity_regularizing_2017}. Increasing the amount of regularization together with the model size is the main approach for improving a neural network without domain-specific knowledge. In the present work, we use a two-layer LSTM Model with a hidden size of 200 for each layer and a final linear layer that reduces the hidden activation count to four. These four activations represent the elements of the estimated attitude quaternion. In order to always generate a unit quaternion, the elements are divided by their Euclidean norm. The structure of the RNN model used in this work is visualized in Figure \ref{fig:model_rnn}.
\begin{figure}
    \centering
    \includegraphics[width=0.45\textwidth,trim={0 -12mm 0 0}]{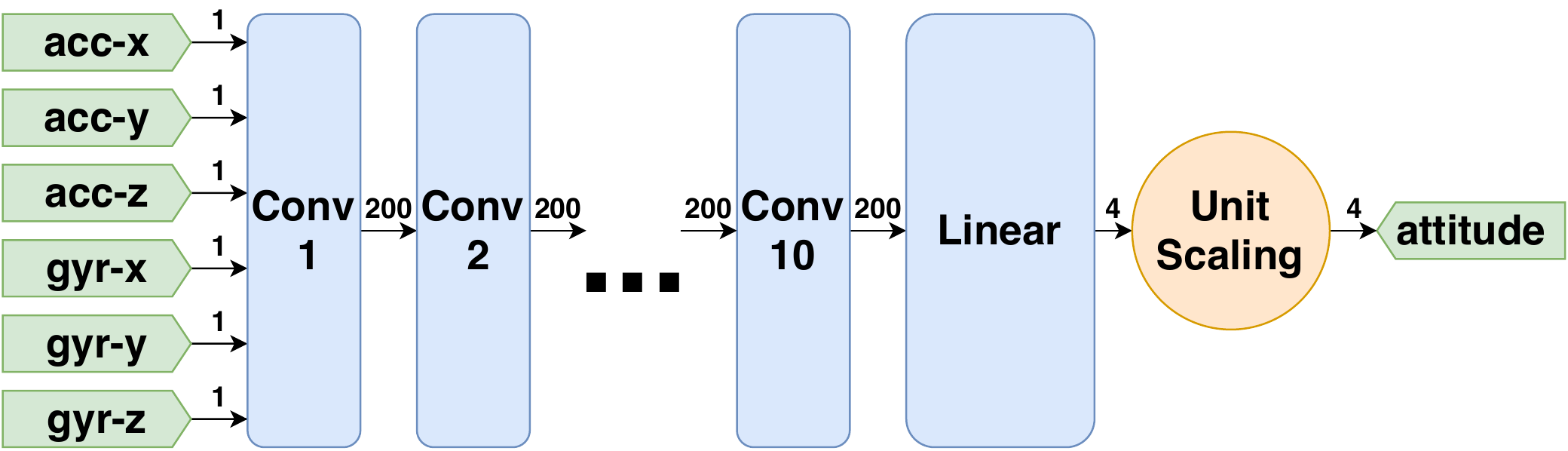}
    \caption{TCN model for attitude estimation. It has the same structure as the RNN model but with 10 convolutional layers instead of 2 RNN layers.}
    \label{fig:model_tcn}
\end{figure}
An alternative approach to RNNs for sequential data are TCNs. TCNs are causal one-dimensional dilated convolutional neural networks with receptive fields big enough to model the system dynamics\cite{andersson_deep_2019}. The main advantage of TCNs compared to RNNs is their pure feed-forward nature. Having no sequential dependencies leads to parallelizability and therefore fast training on hardware accelerators \cite{oord_wavenet:_2016}. The TCN's receptive field describes the number of samples taken into account for predicting a sample. Because TCNs are stateless, the receptive field needs to be large enough to implicitly estimate the system state from the input signals. Because of the dilated convolutional layers, the receptive field grows exponentially with the depth of the neural network allowing for large windows using a manageable amount of layers. In the present work, we use a 10-layer TCN with a receptive field of $2^{10}=1024$ samples and a hidden size of 200 for each layer. The structure of that TCN model is visualized in Figure \ref{fig:model_tcn}.

For linear and convolutional layers, batchnorm \cite{ioffe_batch_2015} is used. Batchnorm standardizes the layer activations, enabling larger learning rates and better generalization. Instead of the commonly used sigmoid or rectified linear unit activation functions, we use Mish, which achieved state-of-the-art results in multiple domains \cite{misra_mish:_2019}. Mish combines the advantages of both activation functions. On the one hand, it is unbounded in positive direction and thus avoids saturation like rectified linear units. On the other hand, it is smooth like sigmoid functions, which improves gradient-based optimization.

For training, long overlapping sequences get extracted from the measured sequences, so the Neural Networks initializes with different states. Because RNNs can only be reasonably trained with a limited amount of time steps for every minibatch, truncated backpropagation through time is used \cite{tallec_unbiasing_2017}. That means that the long sequence gets split in shorter windows that are used for training, transferring the hidden state of the RNN between every minibatch. The measured sequences are standardized with the same mean and standard deviation values to improve training stability \cite{ioffe_batch_2015}.

The main component of the training process is the optimizer. We use a combination of RAdam and Lookahead, which has proven to be effective at several tasks \cite{liu_variance_2019}, \cite{zhang_lookahead_2019}. For the training process, we used the Fastai 2 API, which is built upon Pytorch \cite{howard_fastai_2020}. One of the most important hyperparameters for training a neural network is the learning rate of the optimizer. We choose the maximum learning rate with the learning rate finder heuristic \cite{smith_cyclical_2017} and use cosine annealing for faster convergence \cite{loshchilov_sgdr_2017}. The learning rate finder heuristic determines the maximum learning rate by exponentially increasing the learning rate in a dummy training and finding the point at which the loss has the steepest gradient. Cosine annealing starts with the maximum learning rate, keeps it constant for a given amount of epochs and then exponentially decreases it over time.

The other hyperparameters of the neural network model, such as activation dropout and weight dropout, form a vast optimization space. To find a well-performing configuration, we use population-based training \cite{jaderberg_population_2017}. It is an evolutionary hyperparameter optimization algorithm that is parallelizable and computationally efficient. It creates a population of neural networks with different hyperparameters and trains them for some epochs. Then the hyperparameters and weights of the best performing models are overriding the worst ones, and minor hyperparameter variations are introduced. Repeating this process quickly yields a well-performing solution.

\subsection{Loss Function}
\begin{figure}
    \centering
    \subfigure[Function Comparison]{\includegraphics[width=0.22\textwidth,trim={5mm 0 5mm 10mm}]{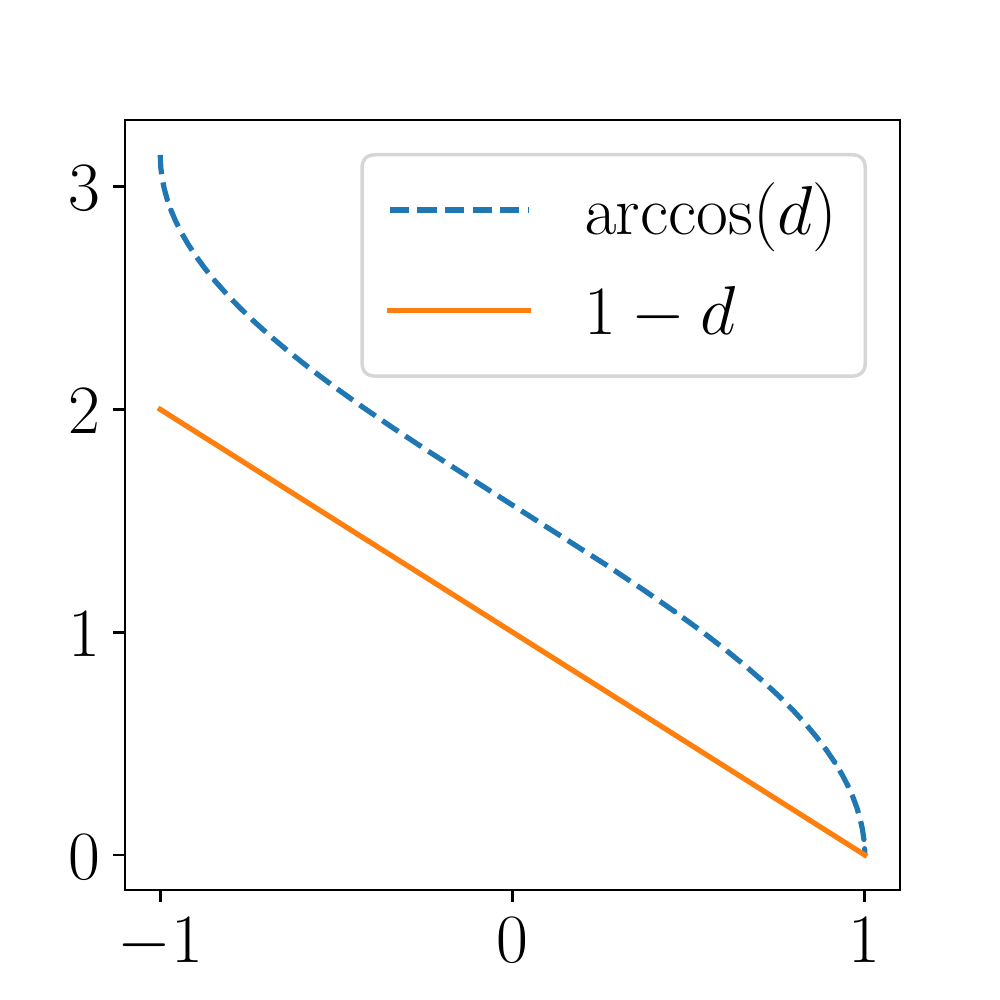}}
    \subfigure[Gradient Comparison]{\includegraphics[width=0.22\textwidth,trim={5mm 0 5mm 10mm}]{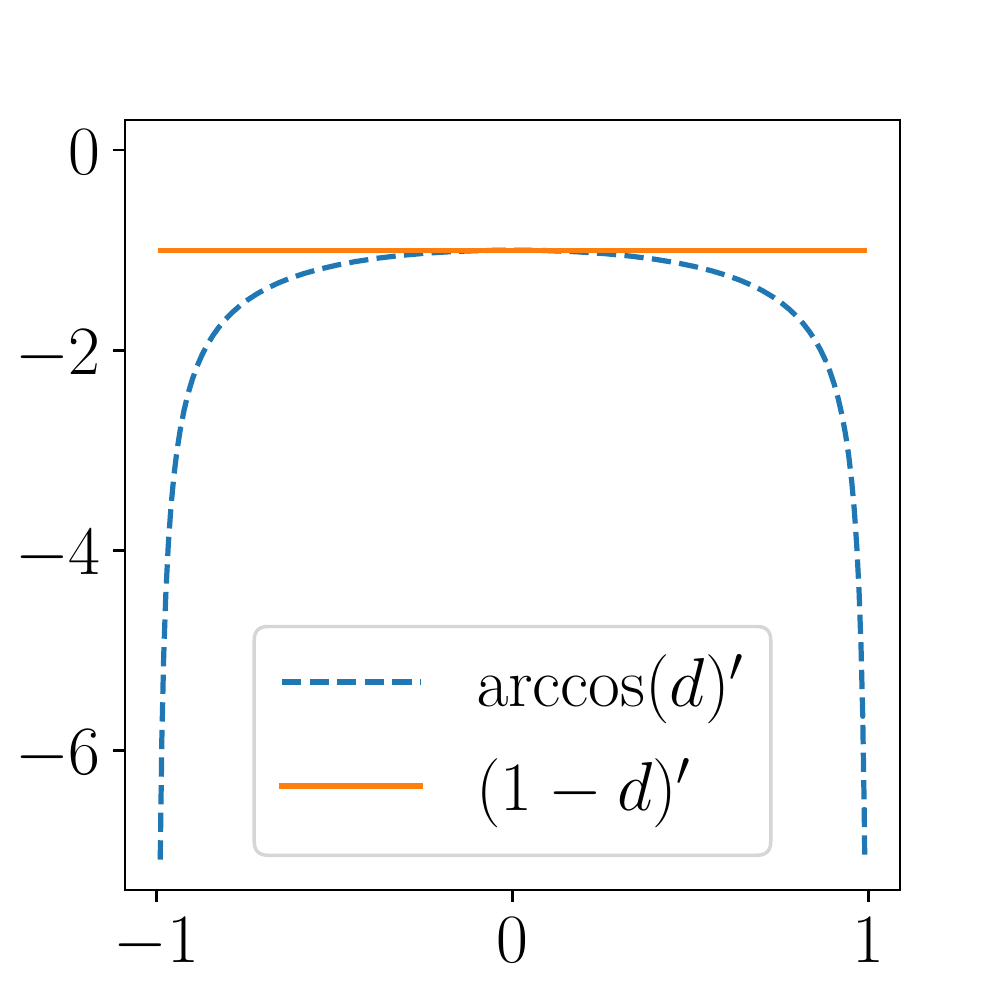}}
    \caption{Comparison of the values and gradients of $\arccos(d)$ and $1-d$. Exploding gradients can be avoided if $\arccos(d)$ is replaced by $1-d$.}
    \label{fig:grad_comp}
\end{figure}

The output of the model is a quaternion that describes the attitude of the sensor. The loss function describes the accumulated error between the estimated and the ground truth values. In most cases, the mean-squared-error between the estimated and reference values are taken. In the present case, an elementwise mean-squared-error of the quaternion is not a reasonable choice, since the orientation cannot be estimated unambiguously with only accelerometer and gyroscope signals --a magnetometer would be necessary. 
An obvious solution would be to choose the loss function equal to the attitude error function $e_\alpha(\mathbf{q}, \mathbf{\hat{q}})$ with

\begin{align}    
  \left[w_\text{err},x_\text{err},y_\text{err},z_\text{err}\right]^\intercal &= \mathbf{q}\otimes\left(\mathbf{\hat{q}}\right)^{-1},\\
  d &= \sqrt{w_\text{err}^2+z_\text{err}^2}, \\
    e_\alpha(\mathbf{q}, \mathbf{\hat{q}}) &=  2\arccos\left(d\right).
\end{align}

However, experiments show that using this error definition leads to unstable training resulting from an exploding-gradient problem. This is caused by the $\arccos$ function, whose derivative function explodes for arguments approaching 1, which is the target of the optimization problem: 
\begin{align} 
    \arccos'(d) &= \frac{-1}{\sqrt{1-d^2}}, \\
    \lim\limits_{d \rightarrow 1}{\arccos'(d)}&=-\infty.
\end{align}
Truncating $d$ close to 1 leads to a solution that is numerically stable with rare exceptions. Replacing the $\arccos$ function with a linear term avoids the exploding gradient completely while keeping the monotonicity and correlation with the attitude:
\begin{equation}
e_{\text{opt}}(\mathbf{q}, \mathbf{\hat{q}})=1-d
\label{eq:arccos_replace}
\end{equation}
Figure \ref{fig:grad_comp} visualizes the differences between both functions and their gradients.

\begin{figure*}
    \centering
    \subfigure[Linear Slow Nonstop]{\includegraphics[width=0.98\textwidth]{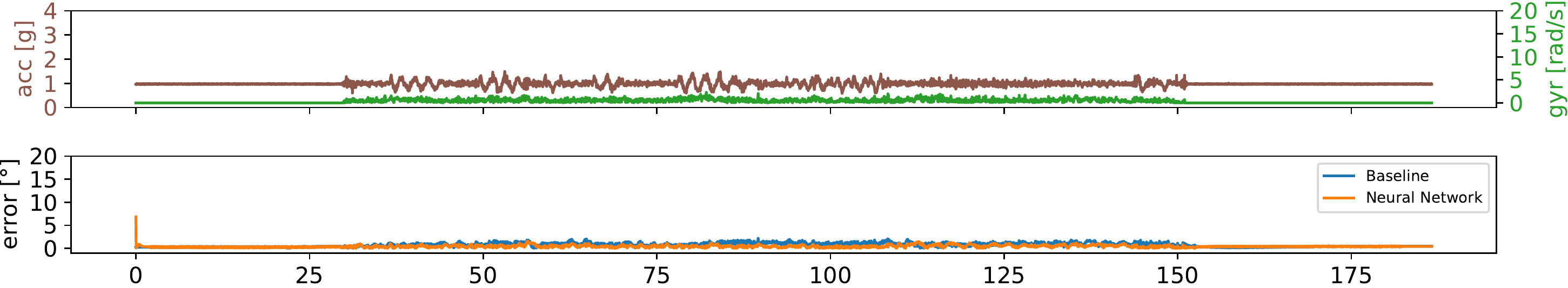}}
    \subfigure[Rotation Fast Paused]{\includegraphics[width=0.98\textwidth]{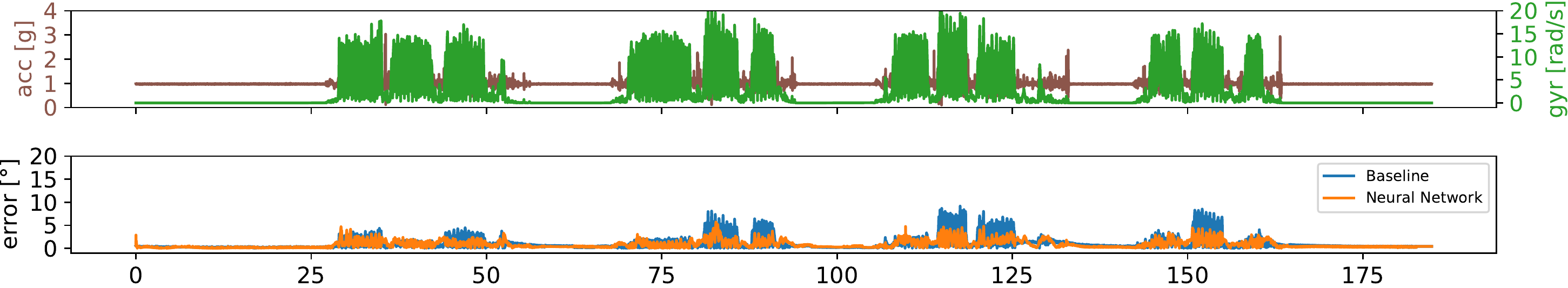}}
    \subfigure[Arbitrary Fast Paused]{\includegraphics[width=0.98\textwidth]{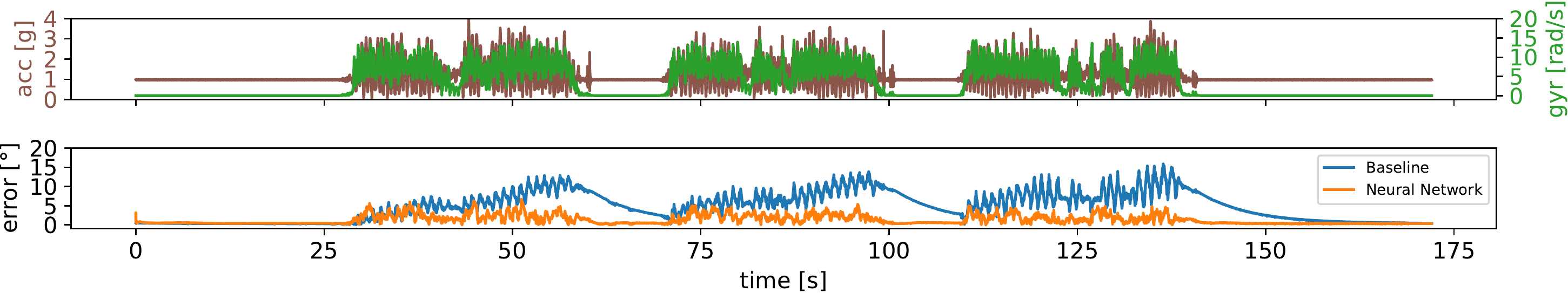}}
    \caption{IMU signal and attitude error comparison for three different motions with different characteristics. In contrast to the baseline filter, the neural network achieves consistently small attitude errors for (a) slow linear motions, (b) fast rotations, and (c) and fast arbitrary motions. More motions in Fig.~\ref{fig:rmse_box_comp}.}
    \label{fig:exampleDataOverTime}
\end{figure*}
Another difficulty of many datasets is the presence of outliers that result from measurement errors. Therefore, we use the smooth-l1-loss function, which is less prone to outliers than the mean-squared-error \cite{feng_wing_2018}.

\subsection{Data Augmentation}
Data augmentation is a method for increasing the size of a given dataset by introducing domain-knowledge. This is a regularization method that improves the generalizability of a model and has already been applied successfully in computer vision\cite{perez_effectiveness_2017} and audio modelling \cite{xiaodong_cui_data_2015}. In case of the present attitude estimation task, we virtually rotate the IMU by transforming the measured accelerometer, gyroscope and reference attitude data by a randomly generated unit quaternion. Thereby, orientation invariance for sensor measurements will be introduced to the model and its susceptibility to overfitting is reduced.

\subsection{Grouped Input Channels}
\begin{figure}
    \centering
    \includegraphics[width=0.45\textwidth]{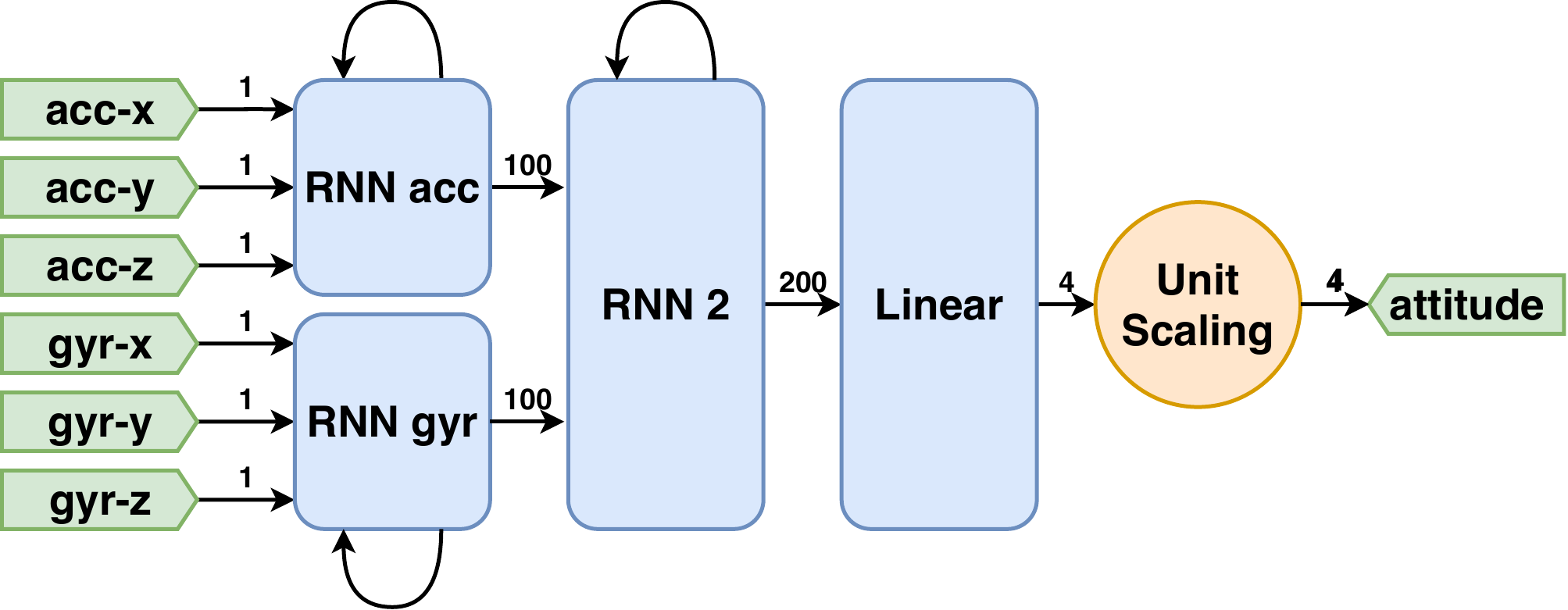}
    \caption{Grouped input version of the RNN model in Figure \ref{fig:model_rnn}. The first RNN layer is split among the accelerometer and the gyroscope signals.}
    \label{fig:grouped_input}
\end{figure}

The default way of processing a multivariate time series is to put all the input signals into the same layer. An alternative way is to create groups of signals that interact with each other and disconnect them from those they don't need to interact with. The idea is to alleviate the neural network's effort in finding interactions between signals. This method has been applied previously to other tasks but without analysis of its impact on the performance \cite{zheng_time_2014}\cite{esfahani_aboldeepio_2019}. In the present application, the accelerometer and gyroscope are grouped separately, with the accelerometer providing attitude information at large time scales and the gyroscope providing accurate information on the change of orientation, as visualized in Figure \ref{fig:grouped_input}.

\section{Experiments}
\begin{figure}
    \centering
    \includegraphics[width=0.48\textwidth]{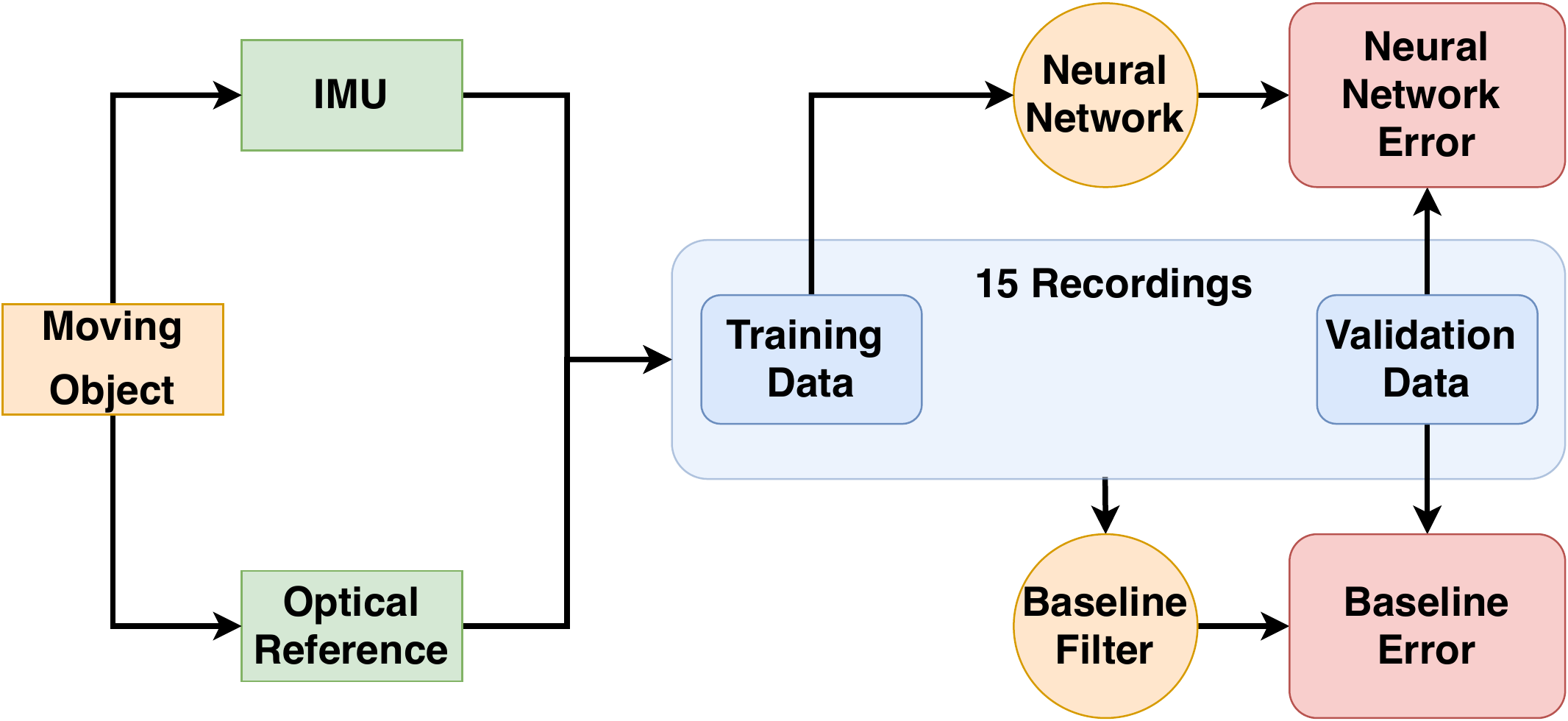}
    \caption{Overview of the performance analysis study with optical ground truth measurements and leave-one-out cross-validation. The performance of the neural network, which is fitted to the training data, and the baseline filter, which is fitted to the entire dataset, is compared for the validation data.}
    \label{fig:experiment}
\end{figure}

The performance of the proposed neural network is compared to the performance of established attitude estimation filters in experiments with a ground truth based on marker-based optical motion tracking. A MEMS-based IMU (aktos-t, Myon AG, Switzerland) is rigidly attached to a 3D-printed X-shaped structure with three reflective markers whose position is tracked at millimeter accuracy by a multi-camera system (OptiTrack, Natural Point Inc., USA). For each moment in time, the three-dimensional marker positions are used to determine a ground-truth sensor orientation with sub-degree accuracy.

To analyze the algorithm performance across different types of motions and different levels of static or dynamic activity, we consider a large number of datasets from different experiments with the following characteristics: 
\begin{itemize}
 \item \texttt{rotation}: The IMU is rotating freely in three-dimensional space while remaining close to the same point in space.
 \item \texttt{translation}: The IMU is translating freely in three-dimensional space while remaining in almost the same orientation.
 \item \texttt{arbitrary}: The IMU is rotating and translating freely in three-dimensional space.
 \item \texttt{slow} versus \texttt{medium} versus \texttt{fast}: The speed of the motion is varied between three different levels.
 \item \texttt{paused} versus \texttt{nonstop}: The motion is paused every thirty seconds and continued after a ten-seconds break or it is performed non-stop for the entire duration of the five-minutes recordings.
\end{itemize}

Different combinations of these characteristics lead to a diverse dataset of 15 recordings each of which contains more than 50,000 samples of accelerometer and gyroscope readings and ground-truth orientation at a sampling rate of 286 Hz. Figure \ref{fig:exampleDataOverTime} shows the Euclidean norms of the three axis of acceleration (acc) and angular rate signal (gyr) over time for three recordings with different combinations of the described characteristics. 

The experimental data is used to validate and compare the following two attitude estimation algorithms: 
\begin{itemize}
    \item \texttt{Baseline}: a quaternion-based attitude estimation filter with accelerometer-based correction steps and automatic fusion weight adaptation \cite{seel_eliminating_2017}. The filter time constant and weight adaptation gain are numerically optimized to yield the best performance across all datasets.
    \item \texttt{Neural Network}: The proposed neural network is trained on a subset of the available (augmented) datasets and validated on the complementary set of data.
\end{itemize}

The characteristics of applying neural networks to the attitude estimation problem are analyzed in three studies. The first one compares the performance of the optimised neural network with the baseline filter. The second one is an ablation study that quantifies the effect of every optimization and compares the performance of the RNN and TCN model. The last study analyzes the effect of scaling the size of the neural network. The workflow of the performance study is visualized in Figure \ref{fig:experiment}.


\subsection{Performance Analysis}\label{sec:perfAnalysis}
In order to compare the performance of the proposed neural network model with established filters, the 15 recordings will be used for a leave-one-out cross-validation. That means that the model will be trained with 15 recordings and validated on the one that was left out. This leads to an increase in computation time because for every recording a new independent model has to be trained, but it provides a better view on the generalizability of the model architecture. The neural network used is the RNN with all the proposed optimizations applied.
\begin{figure}
    \centering
    \includegraphics[width=0.48\textwidth]{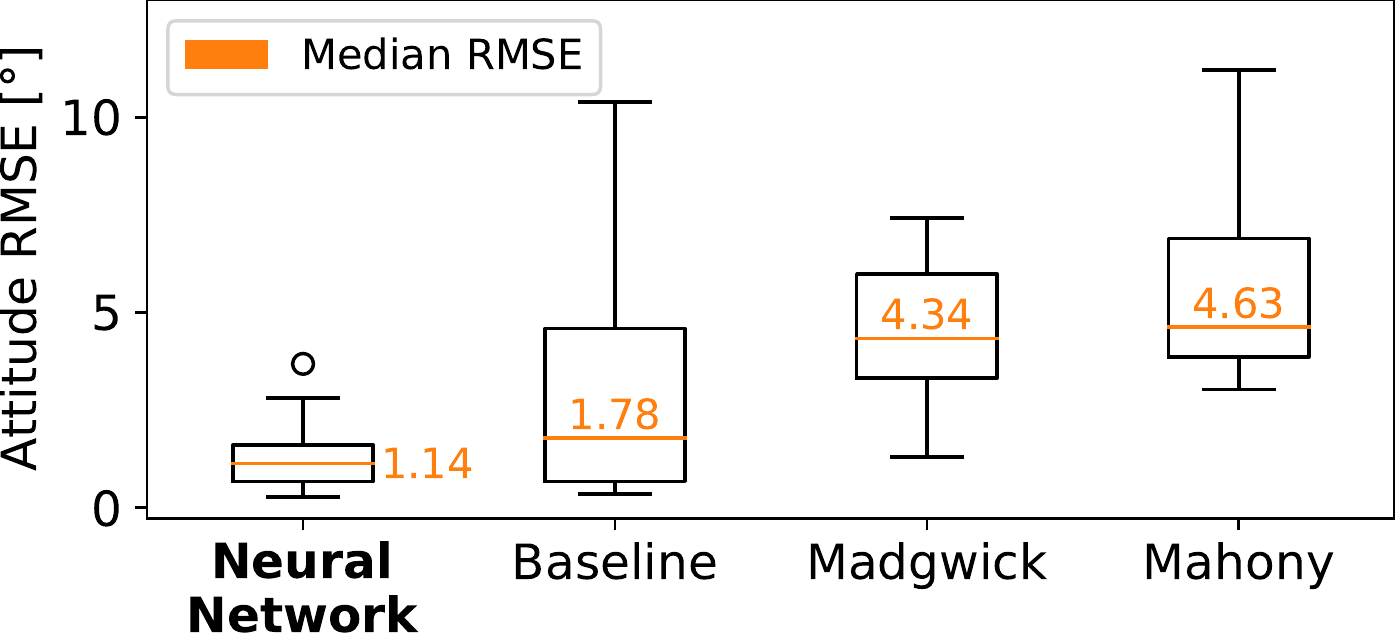}
    \caption{RMSE comparison between the best neural network, the baseline filter~\cite{seel_eliminating_2017}, and two open-source available filter algorithms \cite{noauthor_open_nodate}. Across all types of motions, the proposed neural network achieves clearly smaller median and variance than the conventional filters. Details are presented in Fig.~\ref{fig:rmse_bar_comp}. }
    \label{fig:rmse_box_comp}
\end{figure}
\begin{figure}
    \centering
    \includegraphics[width=0.48\textwidth]{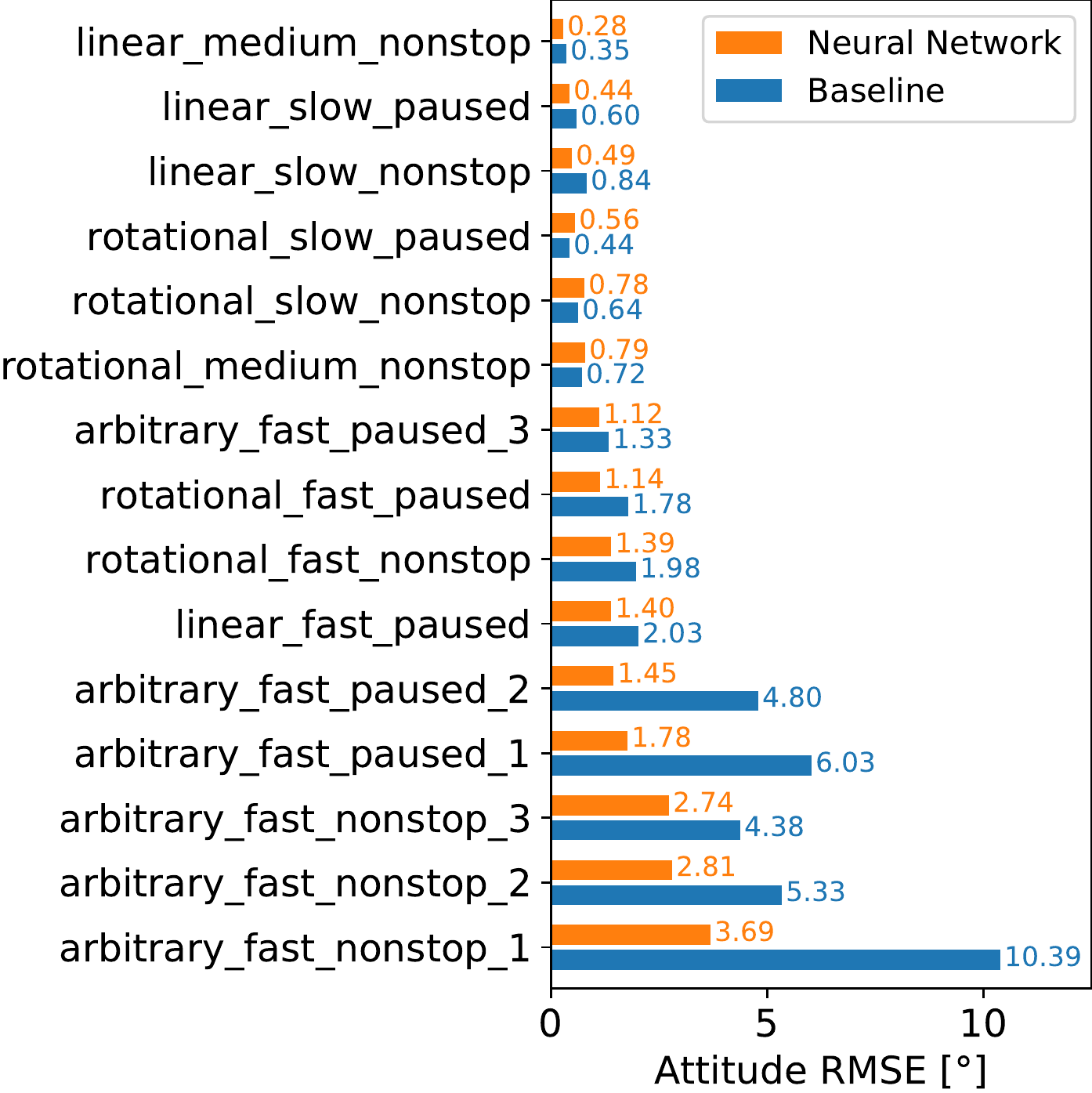}
    \caption{RMSE comparison between the proposed neural network and the best performing conventional filter (baseline) for all different motions.}
    \label{fig:rmse_bar_comp}
\end{figure}

The boxplot in Figure \ref{fig:rmse_box_comp} compares the error distribution of the 15 recordings between the neural network and the baseline. Additionally, two established filters, of which open-source implementations are available \cite{noauthor_open_nodate}, are included. While the baseline filter outperforms the other two filters, the neural network achieves an even smaller median error and performs consistently well across all motions.
The performance comparison between the neural network and the baseline for each recording is visualized in Figure \ref{fig:rmse_bar_comp}. It shows that, in the slow cases, both methods perform similarly, while the baseline filter sometimes diverges in the fast- and arbitrary-motion cases. The diverging behaviour may be observed in Figure \ref{fig:exampleDataOverTime} in the fast arbitrary-motion case. 
Between the movements, when the IMU is resting, the algorithms use the gravitational acceleration to quickly converge towards the true attitude. Overall, the neural network outperforms the established filters significantly, which is even more remarkable in light of the fact that the filters have been optimized on the entire dataset, while the neural network has never seen any of the validation data.

\subsection{Ablation Study}\label{sec:ablStudy}
In the ablation study, the effect of every domain-specific optimization on the performance of the neural network is analyzed. Furthermore, the performance of the RNN and TCN architectures on the attitude estimation problem are compared. In this study, the 15 recordings are split into 12 training recordings and 3 validation recordings. In order to be representative, the validation recordings are the ones that yielded the maximum, minimum and median error in the performance analysis. To both the RNN and TCN architecture with current best practices for time series as basemodels, the three domain-specific optimizations are added iteratively. First, the elementwise mean-squared-error loss is replaced by the optimized attitude error with smooth-l1-loss. In the second step, the data augmentation, which simulates a rotated IMU, is added. In the last step, the input layers are grouped in acceleration and gyroscope signals.

\begin{figure}
    \centering
    \includegraphics[width=0.48\textwidth]{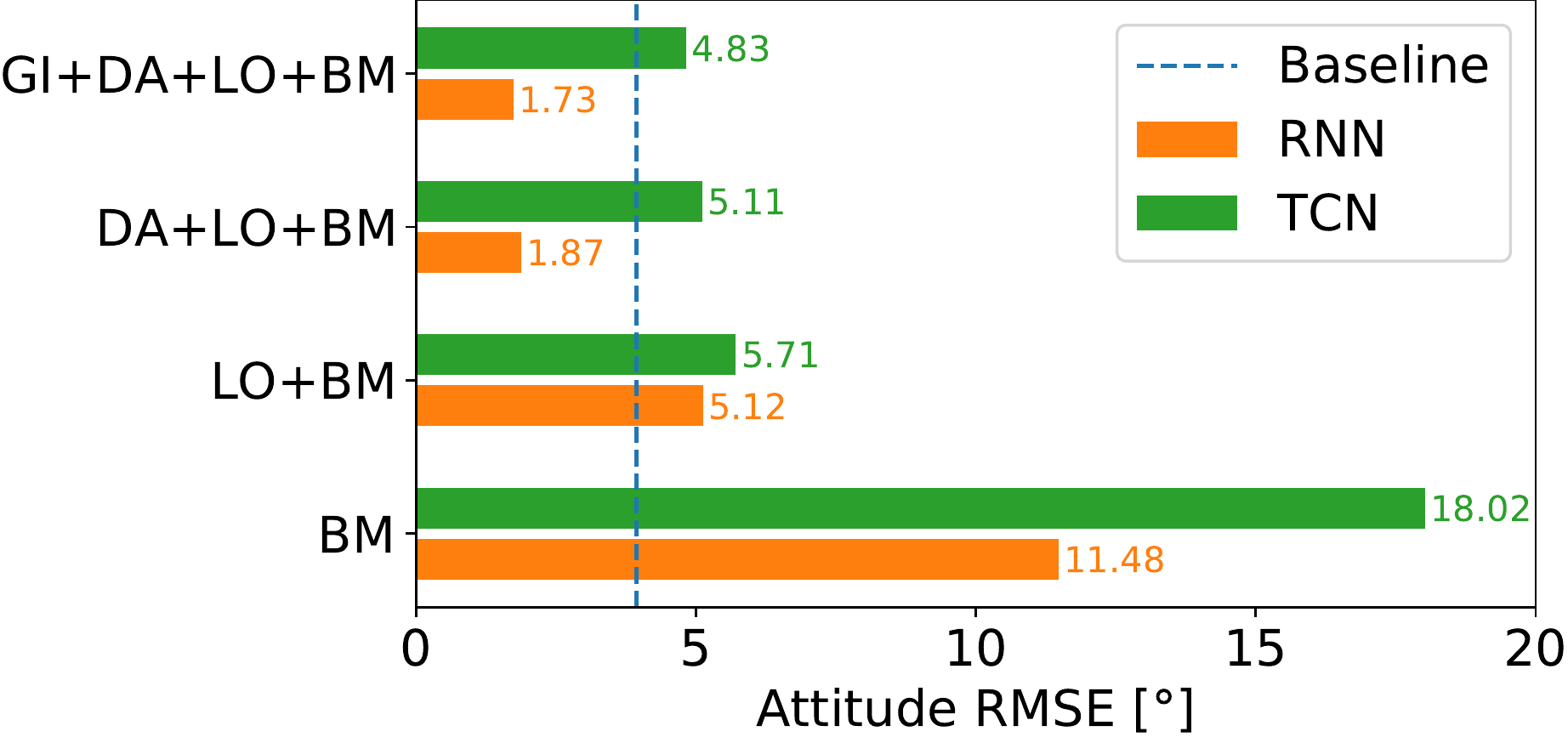}
    \caption{Ablation Study with BM: Basemodel, LO: Loss Optimization, DA: Data Augmentation and GI: Grouped Input. To outperform the baseline filter, domain-specific optimizations of the neural network are required.}
    \label{fig:ablation_study}
\end{figure}

The results of the study are visualized in Figure \ref{fig:ablation_study}. Without the optimizations, the RNN and TCN models perform at a similar level. However, after adding the optimizations, the RNN has a much smaller error. This is plausible with the TCN being limited to its receptive field, while the RNN can track the IMU movement for an indefinite time with its hidden states. Even extending the TCN's receptive field to $2^{14}=16384$ samples, which is a time window of more than $87$ seconds, the results stay the same. When the IMU moves for a longer duration than the time window, the estimation diverges. For this application, especially with real-time applications in mind, the RNN is the better approach.

The second result is that all the optimizations improve both the RNN and the TCN.  Grouping the input leads consistently to minor improvements, while the loss optimization and data augmentation have a significant impact on the performance. When the data augmentation is added to the model, the other general regularization methods need to be reduced or deactivated to avoid over-regularization. Training and validation loss drop with the same pace, which shows that it is very effective at regularizing the model. The same effect probably could be achieved by increasing the size of the dataset by several orders of magnitude, which would require more costly recordings.

The final result is that both the loss optimization and the data augmentation are necessary to outperform the baseline filter. Without these domain-specific optimizations, even the highly optimized general purpose neural networks do \emph{not} generalize well enough. If all aforementioned optimizations are applied, the neural network performs significantly better than the baseline filter.

\subsection{Model Size Analysis}
In order to analyze the effect of the model size to the attitude error, the RNN model from the performance analysis (Section~\ref{sec:perfAnalysis}) is applied to the 12 training and 3 validation recordings of the ablation study (Section~\ref{sec:ablStudy}). The number of neurons of each layer of the RNN is scaled from 10 to 200, and the attitude error is compared.
\begin{figure}
    \centering
    \includegraphics[width=0.45\textwidth]{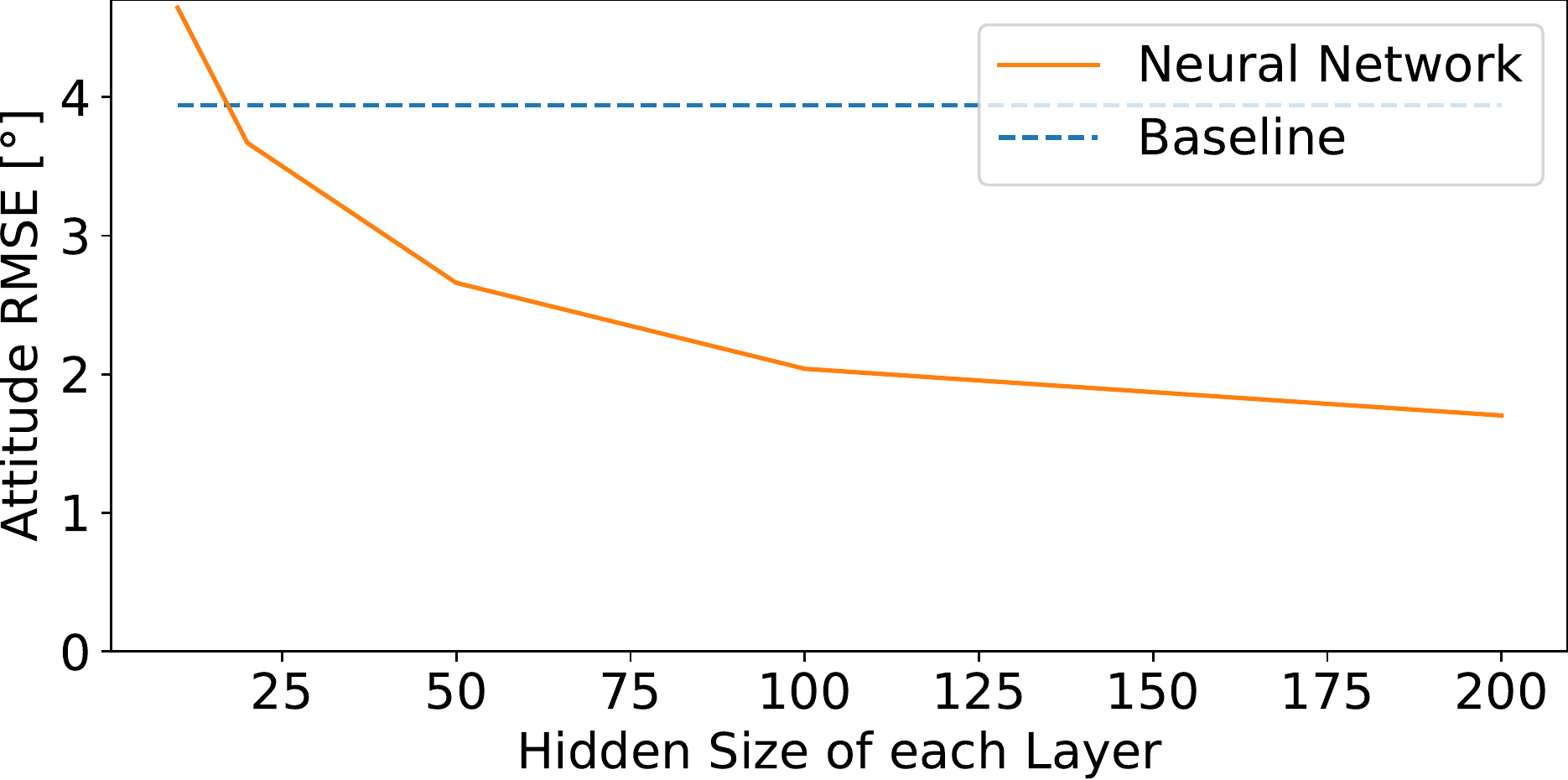}
    \caption{Model Size Analysis. A network with 25 neurons per layer outperforms the baseline.}
    \label{fig:size_study}
\end{figure}
The results of the study are visualized in Figure \ref{fig:size_study}. As expected, the error decreases with increasing hidden size, with the gradient decreasing at bigger neuron counts. In this example, 20 neurons per layer are already enough to achieve the same mean attitude error as the baseline filter. Decreasing the hidden size of RNNs helps to reduce the memory footprint and overall computation time, which is important for embedded systems. But it only marginally reduces the training and prediction time on hardware accelerators with high parallelization capabilities, because of its sequential nature.


\section{Conclusion}
This work has shown that neural networks are a potent tool for IMU-based real-time attitude estimation. If domain-specific optimizations are in place, then large recurrent neural networks can outperform state-of-the-art nonlinear attitude estimation filters. These optimizations require knowledge about the process that the neural network identifies. However, it does not require specific knowledge (equations, signal characteristics, parameters) that is needed for implementing a well-performing filter. Another requirement for the neural-networks-based solution is a sufficiently rich set of data with ground truth attitude. However, data augmentation was proven to reduce this demand significantly. Finally, depending on the number of parameters, more computation time is needed for the attitude estimation with neural networks.

Leave-one-out cross-validation was used to show that the trained network performs well on new data from motions that were \emph{not} used for training. Networks trained with a sufficiently broad range of motions can be applied to arbitrary tasks that use the same IMU and sampling rate. Future research will focus on generalizing the model to data from different IMUs with different sampling rates and different error characteristics. This will answer the question whether a sufficiently trained neural network can be used as a competitive solution in new sensor and environment settings without the need for collecting and using new training data.

\section*{Funding}
This work was partly funded by the German Federal Ministry of Education and Research (FKZ: 16EMO0262).

\bibliographystyle{./bibliography/IEEEtran}

\end{document}